\documentclass[letterpaper, 12pt]{article}

\usepackage{algorithm}
\usepackage{algpseudocode}
\usepackage{amsfonts}
\usepackage{amsmath}
\usepackage{amssymb}
\usepackage{array}
\usepackage{geometry}
\usepackage{graphicx}
\usepackage[colorlinks=true, citecolor=blue, filecolor=black, linkcolor=red, urlcolor=blue]{hyperref}
\usepackage{textcomp}
\usepackage{booktabs}
\usepackage{array}
\usepackage{url}
\usepackage{color}
\usepackage{nomencl}

\title{Stochastic Deep Koopman Model for Quality Propagation Analysis in Multistage Manufacturing Systems}

\author{Zhiyi Chen\thanks{Department of Mechanical Engineering, University of Michigan, Ann Arbor, MI, USA
        {\tt\small \{chzhiyi, cohenyo, xhuan, xingjian, junni\}@umich.edu} }
        \and 
        Harshal Maske\thanks{Ford Motor Company, Dearborn, MI, USA
        {\tt\small hmaske1@ford.com, huanyis@umich.edu, deveshu@gmail.com, mhopka@gmail.com } }
        \and
        Huanyi Shui\footnotemark[2]
        \and
        Devesh Upadhyay\footnotemark[2]
        \and
        Michael Hopka\footnotemark[2]
        \and
        Joseph Cohen\footnotemark[1]
        \and
        Xingjian Lai\footnotemark[1]
        \and
        Xun Huan\footnotemark[1]
        \and
        Jun Ni\footnotemark[1]
}

\date{}

\begin{document}

\maketitle

\begin{abstract}
  The modeling of multistage manufacturing systems (MMSs) has attracted increased attention from both academia and industry. Recent advancements in deep learning methods provide an opportunity to accomplish this task with reduced cost and expertise. This study introduces a stochastic deep Koopman (SDK) framework to model the complex behavior of MMSs. Specifically, we present a novel application of Koopman operators to propagate critical quality information extracted by variational autoencoders. Through this framework, we can effectively capture the general nonlinear evolution of product quality using a transferred linear representation, thus enhancing the interpretability of the data-driven model. To evaluate the performance of the SDK framework, we carried out a comparative study on an open-source dataset. The main findings of this paper are as follows. Our results indicate that SDK surpasses other popular data-driven models in accuracy when predicting stagewise product quality within the MMS. Furthermore, the unique linear propagation property in the stochastic latent space of SDK enables traceability for quality evolution throughout the process, thereby facilitating the design of root cause analysis schemes. Notably, the proposed framework requires minimal knowledge of the underlying physics of production lines. It serves as a virtual metrology tool that can be applied to various MMSs, contributing to the ultimate goal of Zero Defect Manufacturing.
\end{abstract}

\section{Introduction}
\label{sec: introduction}
Quality control is critical for ensuring the consistency of product quality in the modern manufacturing industry. Improving product quality not only leads to increased customer satisfaction and company profitability but also contributes to sustainability by reducing material waste. Recognizing the growing industrial demands, Zero-defect Manufacturing (ZDM) has gained increased attention in recent years \cite{psarommatis2020zero}. ZDM proposes four strategies, namely detect, predict, repair, and prevent, encompassing the entire product manufacturing lifecycle, enabling continuous quality improvements \cite{psarommatis2022zero}. By proactively identifying defects at early stages, ZDM aims to prevent quality disruptions and reduce costs. Within the framework of ZDM, defect detection and prediction play fundamental roles, as they provide operators with crucial information to plan for repair and preventive actions.

ZDM encourages the use of advanced technologies to improve manufacturing processes. Traditional quality control schemes (e.g., statistical process control or SPC) rely on physical inspections and control charts to monitor processes and alert operators when anomalies are detected \cite{yeh2006multivariate,lowry1995review}. However, such inspections are limited to specific stations and cannot cover every product, posing challenges in comprehensive quality assessment \cite{mandroli2006survey}. Recent advancements in sensing and automation technology have demonstrated that improved data collection and in-situ analytics enable the development of innovative virtual metrology (VM) models \cite{shui2018twofold}. These VM models are developed offline using historical data and then deployed online to provide accurate estimations of product quality. Therefore, the VM method is categorized as a product-oriented approach in ZDM. The deployment of VM models holds great value as they provide guidance to enhance manufacturing systems.

Quality estimation using physics-based VM models is appealing due to the models' interpretability. However, obtaining an accurate physical model can be challenging and costly, especially for high-complexity manufacturing operations. Furthermore, unknown parameter values, uncertain assumptions from physics, and environmental noise all contribute to model inaccuracies \cite{yao2020cooperative}. As a result, researchers have explored data-driven methods to fill these gaps \cite{wang2022hybrid}. For example, Bai \emph{et al.} compared different feature engineering techniques and used a support vector machine (SVM) to predict quality indices using features extracted from process measurements \cite{bai2019comparison}. However, such conventional machine learning approaches require expert knowledge to conduct feature engineering, rendering them impractical for tasks with high-dimensional and nonintuitive data. In this study, we aim to develop a data-driven model that can automatically extract valuable features from process measurements and provide accurate quality estimations for complex manufacturing systems. This model serves as a VM tool, allowing for real-time monitoring of product quality and aligning with the overarching objective of ZDM.
\section{Literature Review}
\label{sec: literature_review}
Researchers have significantly progressed in developing the conceptual framework to integrate VM with manufacturing systems. A systematic review of the VM methods was found in \cite{dreyfus2022virtual}. Among those, variants of deep neural network (DNN) architectures have emerged as effective approximators for complex systems with high-dimensional data. Yuan \emph{et al.} proposed an augmented long short-term memory (LSTM) network to learn the quality-relevant hidden dynamics from long-term sequential data \cite{yuan2019nonlinear}. When both time-sequential and time-invariant data were involved, Ren \emph{et al.} combined LSTM with an improved Wide-and-Deep model to capture the key quality information \cite{ren2020wide}. When image data was involved, convolutional neural networks (CNNs) were used to pre-process and detect defects from various processes, including welding and additive manufacturing \cite{miao2022real, li2020quality}. These approaches effectively delegate feature engineering to DNNs and sidestep the need for human expertise-based feature design. However, a key drawback of DNN models is their poor interpretability. More importantly, the above studies are all limited to single-stage processes.

Dreyfus \emph{et al.} emphasized in their recent review paper that there was a lack of research addressing VM models for multistage manufacturing systems (MMSs) \cite{dreyfus2022virtual}. It is important to note that most real-life production lines operate as MMSs, involving multiple stations. Generally, modeling MMSs is more challenging than single-stage systems due to confounding interstage couplings. In MMSs, product quality at each stage is a function of the previous steps, the current operation, and stage-specific noise factors. Conventional data-driven methods, e.g., the ensembled tree-based method in \cite{chen2020virtual} and the shallow neural network in \cite{yacob2021multilayer}, aggregate information from all stages together to build end-to-end models that do not reveal the coupling between stages. Consequentially, the predictions may be overwhelmingly opaque for operators to interpret. Identifying the problematic operation can become a bottleneck in such cases, leading to significant delays in introducing corrections \cite{liu2010variation}. Therefore, methods that can capture the interstage dynamics while enabling straightforward interpretation are required.

Filz \emph{et al.} suggested that monitoring the state of intermediate products could provide valuable insights into the interstage behavior in MMS \cite{filz2020data}. They proposed using clustering methods to categorize different classes of intermediate products. Arif \emph{et al.} combined multiple principal component analysis (PCA) modules with decision trees for binary classification of products \cite{arif2013cascade}. Liu \emph{et al.} developed an algorithm that learned the classification model through three steps: Quality embedding, Temporal-interactive learning, and Decoding (QTD) \cite{liu2020adversarial}. Notably, these approaches do not involve direct prediction of product quality.  Zhang \emph{et al.} proposed a path enhanced bidirectional graph attention network (PGAT) to learn the long-distance machine dependencies \cite{zhang2021path}. Wang \emph{et al.} introduced a multi-task joint learning approach and designed a loss function to automatically learn the importance of different operation stages and quality indices, leading to improved prediction performance \cite{wang2023production}. Such approaches were developed to enable quantitative prediction of quality while paying more attention to learn the interaction between stages. However, how to leverage those frameworks for supporting operational decisions remains an open question. In contrast, Bayesian methods in \cite{papananias2019bayesian} and \cite{mondal2021monitoring} could describe the uncertainty associated with different operations and aid in diagnosing root causes. Nevertheless, these methods often require prior knowledge of the distributions of process variables, making them less practical in complicated MMSs. Zhao \emph{et al.} combined LSTM and genetic algorithm to enhance quality control for a multistage boring process \cite{zhao2023novel}. Yet, the model cannot provide in-process quality estimations for intermediate products, leading to less flexibility in conducting process improvements. 

Stream-of-variation (SoV) analysis is the mainstream methodology widely implemented for modeling MMSs (see \cite{liu2010variation, shi2009quality} for comprehensive reviews). SoV uses state space representations to capture the propagation of quality variations and predict stage-by-stage product quality in MMSs to reduce variation. However, using SoV requires linear dynamics, and therefore, these techniques have been mostly limited to machining and assembly processes. Despite these shortcomings, SoV analysis has demonstrated the benefits of modeling quality propagation; namely, it provides better visibility of a process and greatly helps with implementing predictive control schemes \cite{djurdjanovic2007online}. Following this idea, Shui \emph{et al.} proposed a hybrid approach that integrated data-driven methods into SoV to model roll-to-roll processes \cite{shui2018twofold}. Nevertheless, the proposed hybrid model relied on manual feature selection, which still presented a challenge for large-scale systems. In \cite{jiang2014real}, Jiang \emph{et al.} designed a directed graph to represent the topology of multistage machining processes, then used neural networks (NNs) to encode the quality evolution between different stations. The method involved training multiple distinct NNs, possibly leading to cumulative prediction errors. Yan \emph{et al.} designed a multi-task stacked DNN to predict all sensing outputs jointly, and used a two-layer neural network to model the quality propagation between stages \cite{yan2021deep}. However, the method yielded a model with nonlinear transitions between stages, making it difficult to design algorithms (e.g., process control \cite{djurdjanovic2007online}, system design \cite{abellan2011design} and tolerance allocation \cite{ding2005process}) to improve the production system. Recently, Lee \emph{et al.} introduced the concept of Stream-of-Quality (SoQ) as an extension of SoV, enabling the generalization of SoV principles to a broader range of nonlinear systems \cite{lee2022stream}. This advancement offered new opportunities to integrate quality propagation analysis into VM techniques.

We propose a new framework to model the quality propagation in MMSs that possesses three key desirable properties: (1) automated processing of sensor data, enabling scalability for large-scale manufacturing systems; (2) extensibility for nonlinear systems; and (3) interpretability to justify reasoning facilitated by a linear latent dynamics modeling, supporting decision-making to improve the MMS. Specifically, the linear dynamics in (3) are enabled by a stochastic deep Koopman (SDK) framework that models the complex behavior of MMSs in a transformed linear latent space and predicts product quality on a per-stage basis. Table \ref{table1} summarizes the advancements of the proposed method.

\begin{table}
\scriptsize
\centering
\setlength{\extrarowheight}{1pt}
\caption{Comparison between the Relevant Studies}
\label{table1}
\setlength{\tabcolsep}{1pt}
\begin{tabular}{p{0.2\linewidth}  p{0.2\linewidth}  p{0.15\linewidth}  p{0.16\linewidth}  p{0.12\linewidth}  p{0.12\linewidth}}
\toprule
\textbf{Methods} & \textbf{Automated Feature\newline Engineering}    & \textbf{Model Type}   & \textbf{Demonstrate\newline Interpretability} & \textbf{Stagewise\newline Prediction}& \textbf{Stochasticity} \\
\midrule
Arif, 2013 \cite{arif2013cascade}               &   No    &   Nonlinear   &    Yes  &   Yes  &   No \\
Liu, 2020 \cite{liu2020adversarial}             &   Yes   &   Nonlinear   &    No   &   No   &   No \\
Zhang, 2021 \cite{zhang2021path}                &   Yes   &   Nonlinear   &    Yes  &   No   &   No \\
Wang, 2023 \cite{wang2023production}            &   Yes   &   Nonlinear   &    No   &   Yes  &   No \\
Zhao, 2023 \cite{zhao2023novel}                 &   Yes   &   Nonlinear   &    No   &   Yes  &   No \\
Papananias, 2019 \cite{papananias2019bayesian}  &   No    &   Linear      &    Yes  &   No   &   Yes\\
Mondal, 2021 \cite{mondal2021monitoring}        &   No    &   Nonlinear   &    Yes  &   Yes  &   Yes\\
Shui, 2018 \cite{shui2018twofold}               &   No    &   Linear      &    Yes  &   No   &   No\\
Jiang, 2014 \cite{jiang2014real}                &   No    &   Nonlinear   &    No   &   Yes  &   No\\
Yan, 2021 \cite{yan2021deep}                    &   Yes   &   Nonlinear   &    No   &   Yes  &   No\\
\textbf{SDK from this work}                               &   \textbf{Yes}   &   \textbf{Nonlinear}   &    \textbf{Yes}  &   \textbf{Yes}  &   \textbf{Yes}\\

\bottomrule

\end{tabular}
\end{table}

\begin{figure}
\centerline{\includegraphics[width=\columnwidth]{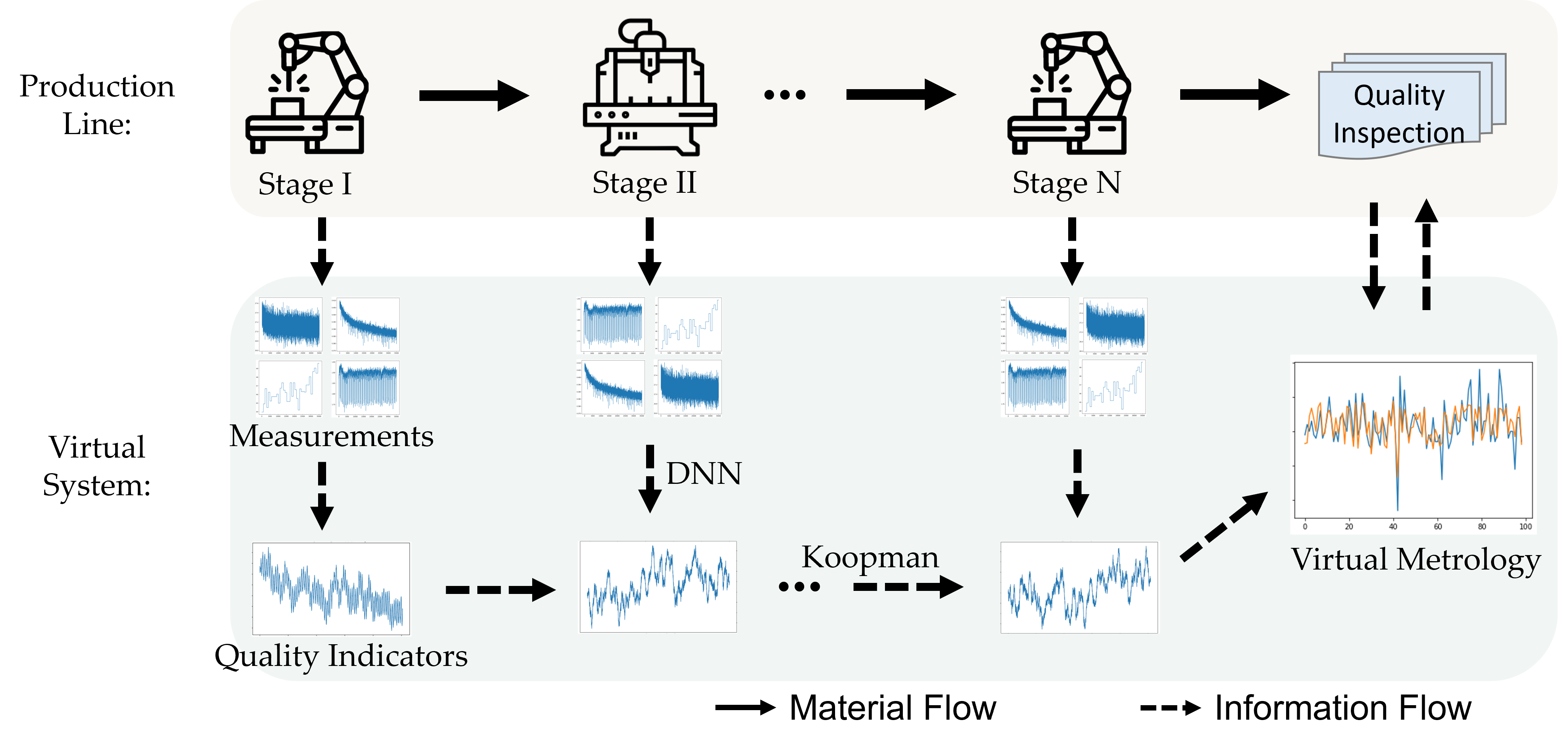}}
\caption{Overview of the proposed Deep Koopman framework.}
\label{fig:fig1}
\end{figure}

An overview of the methodology is illustrated in Fig. \ref{fig:fig1}. We implement a variational autoencoder (VAE) for each stage to translate process data into a set of quality indicators, serving as a latent expression of the crucial quality information. The evolution of the quality indicators along a production line is then captured by Koopman transition models. We also introduce auxiliary networks to simplify the transition models into an interpretable structure while retaining the dynamics on the intrinsic, embedded latent coordinates. The novel combination of VAEs and Koopman operators maps a nonlinear MMS onto a Hilbert space where the state evolution is approximately linear. The transformed linear model can be regarded as a new variant of SoV models. Therefore, the range of existing SoV-based techniques that require linear dynamics, such as process optimization and tolerance allocation, remain usable under this formulation. This is in contrast to the aforementioned nonlinear modeling approaches in literature that can no longer take advantage of these SoV tools. 

The main contributions of this work are as follows.
\begin{enumerate}
\item \emph{Virtual metrology}: A VM framework is specifically designed for MMSs. The method differs from conventional VM methods because it emphasizes capturing quality propagation in MMSs.
\item \emph{Linearity}: Koopman operators are devised to capture nonlinear quality propagations through linear representations. This provides better process visibility and interpretability to facilitate locating the root causes of potential quality variations.
\item \emph{Stochasticity}: A novel stochastic deep learning technique is introduced to map raw data onto a unified feature space. By training the encoders to remove redundant process information, the model achieves enhanced robustness to process and sensor noises.
\end{enumerate}

This paper is organized as follows. Sec. \ref{sec: problem statement} describes the overall problem statement. Sec. \ref{sec: Koopman} then presents a preliminary background on the Koopman operator, followed by the overall SDK methodology in Sec. \ref{sec: proposed method}. The proposed method is then tested on a public dataset in Sec. \ref{sec: case study}. Finally, Sec. \ref{sec: conclusion} provides concluding remarks and future work.
\section{Problem Statement}
\label{sec: problem statement}

Consider an archetypal $N$-stage production line with $N \geqslant 2$. Each stage is denoted by $S_k$, $k=1, 2, \ldots, N$, and has $m_k$ machines. Each machine at the $k^{\rm th}$ stage is denoted by $M_{k,i}$, $i=1, 2, \ldots, m_k$. Let $p_{k,i}$ be the number of process measurements taken from machine $M_{k,i}$, and the total number of process measurements from stage $S_k$ is $p_k= \sum_{i}{p_{k,i}}$. The process measurements include material properties, environmental and human factors (which may be considered a shared variable for multiple machines within a stage), and process parameters (which are specific to each machine). We aggregate all process measurements from stage $S_k$ into a single measurement vector of dimension $p_k$. Then, Each process measurement is denoted by $x_{k,i}$, $i=1, 2, \ldots, p_k$. Similarly, the quality indices at $S_k$ are denoted by $y_{k,i}$, $i=1, 2, \ldots, q_k$, where $q_k$ is the total number of quality indices at stage $S_k$.

The quality estimation task is to find mappings $g_k$ so that
\begin{equation} 
\tilde{Y}_k=g_k(X_1, ..., X_k), \qquad k \geq 1 \label{xyCorrelation}
\end{equation}
where $\tilde{Y}_k=[\tilde{y}_{k,1}, \ldots, \tilde{y}_{k,q_k}]^T$ are the estimated quality indices and $X_k=[x_{k,1}, \ldots, x_{k,p_k}]^T$ are the process measurements specific to each product. The structure of Eq. \eqref{xyCorrelation} indicates that quality $Y_k$ is causally affected by $X_1$ through $X_k$, but not by the process variables beyond the $k^{\rm th}$ stage.

To accomplish this task, a conventional SoV-based model can be constructed as follows:
\begin{align} 
\tilde{Y}_i &= A_i \tilde{Y}_{i-1} + B_iX_i, \qquad i=1, 2, \ldots, k  \label{SoVPropagation} \\
\tilde{Y}_k &= \sum_{i=1}^k \big( \Gamma_i X_i \big)
\end{align}
where $A_i, B_i$ and $\Gamma_i$ are transformation matrices. Eq. \eqref{SoVPropagation} indicates that a linearized mapping between $Y$ and $X$ is used to model a generic nonlinear system.
\section{Preliminary: Koopman Operator Theory}
\label{sec: Koopman}

This section briefly introduces the advancements in Koopman operator theory to provide a clear understanding of this work.

The linearization of a nonlinear dynamic system, i.e., $x_{t+1}=F(x_t)$ with $x\in \mathbb{R}^n$ and $t \in \mathbb{N}_0$, is a common practice, but conventional linearized models provide reduced-order approximations with local behavior. In 1931, B.O. Koopman proposed an alternative, global linearization method that maps the original state of a nonlinear system onto an infinite dimensional Hilbert space via possible measurements $\gamma(x_t)$ \cite{koopman1931hamiltonian}. The Koopman operator $\mathcal{K}$ then linearly advances the system dynamics in the invariant subspace as 
\begin{equation} 
\gamma(x_{t+1})=\gamma \circ F(x_t)=\mathcal{K} \gamma(x_t). \label{koopmanEqn}
\end{equation}

In practice, a finite-dimensional approximation of the infinite-dimensional $\mathcal{K}$ is sought, for example, by numerical methods such as the dynamic mode decomposition (DMD) and extended DMD \cite{schmid2010dynamic,williams2015data}. Recently, Lusch \emph{et al.} used autoencoders (AEs) to identify the Koopman eigenfunctions and achieved high accuracy in predicting the behavior of nonlinear fluid flow \cite{lusch2018deep}. The eigenfunctions $\eta(x)$ are designed deliberately, satisfying:
\begin{equation} \eta(x_{t+1})=\mathcal{K}(\lambda) \eta(x_t)=\lambda\eta(x_t) \label{EigenEqn}
\end{equation}
where $\lambda$ represents the corresponding eigenvalues. In this eigenspace, $\mathcal{K}$ can be approximated by a block-diagonal Koopman matrix. Balakrishnan \emph{et al.} further improved the performance of the deep Koopman model by integrating it with a generative adversarial network and enforcing a stochastic latent embedding \cite{balakrishnan2021stochastic}.
While most existing works utilize Koopman models to predict the evolution of time series systems, we will instead use Koopman operators to propagate quality characteristics over stages in an MMS.
\section{Proposed Method}
\label{sec: proposed method}

\begin{figure}
\centerline{\includegraphics[width=\columnwidth]{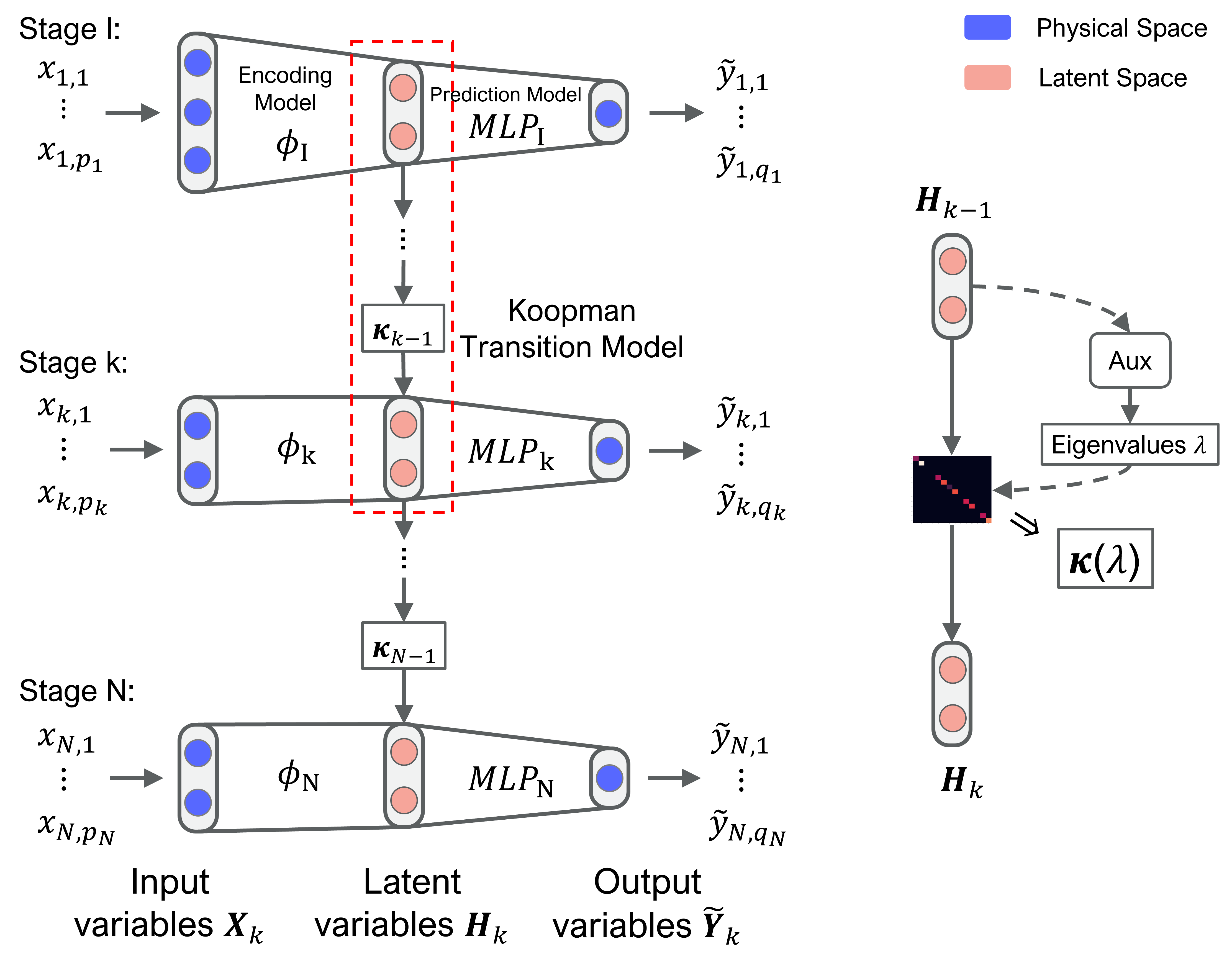}}
\caption{In the base framework, the quality indicators are propagated in a deterministic latent space. The Koopman transition models are constructed from eigenvalues that are parameterized by auxiliary networks.}
\label{fig:fig2}
\end{figure}

We will first present the proposed deep Koopman framework for MMSs through a base model that uses a deterministic expression of the quality indicators in the latent space. Then, an augmentation that enforces a stochastic expression in the latent space will be introduced. We will demonstrate through a case study that the stochastic model outperforms the deterministic model due to its robust capability to handle noise.

Fig. \ref{fig:fig2} illustrates the base model of the proposed framework, consisting of three modules for each stage: the encoding module, the Koopman transition module, and the prediction module. At the $k^{\rm th}$ stage, the model takes process measurements $X_k$ as the inputs, and the encoding model extracts the temporal quality indicators. Combining with the quality information propagated by a Koopman transition model $\mathcal{K}_{k-1}$ from the upstream stage, the latent quality indicators $H_k$ are computed. A multilayer perceptron (MLP) network is designed to serve as the prediction model that predicts the quality indices using the quality indicators. Note that the quality propagation starts from $S_1$ and terminates at $S_N$. 

In contrast to Eq. \eqref{SoVPropagation}, the proposed method brings two benefits: 1) quality propagation is modeled using critical parameters instead of raw process data; and 2) a linear embedding is captured for a nonlinear system instead of applying simple linearization.

\subsection{Encoding module by AE}
\label{sec: AE}
In general, the evolution of quality information in MMSs is nonlinear. We intend to find a suitable invariant subspace where this nonlinear propagation can be approximated well by a linear embedding. Based on Koopman's theory, the linear approximation should converge to the original propagation system as we increase the dimension of the linearly embedded space. This contrasts significantly with standard feature engineering techniques in machine learning, which typically aim to reduce dimensionality with techniques such as PCA. We note that the selection of this subspace is not unique. Since AE allows one to construct flexible mappings from $X_k$, we use it to identify the best invariant subspace where quality indicators simultaneously serve as the eigenfunctions in Eq. \eqref{EigenEqn}.

The encoding module for the $k^{\rm th}$ stage takes process measurements as inputs and returns the temporal quality indicators, as given by:
\begin{equation} \hat{H}_k=\phi_k(X_k) \label{encoder}\end{equation}
where $\hat{H}_k\in\mathbb{R}^{d_{h,k}}$ and $d_{h,k}$ indicates the dimension. Normally, the AE consists of an encoder $\phi_k$ and a decoder $\psi_k$. In \cite{lusch2018deep}, the authors used the encoder to derive the latent variables and the decoder to map from the latent space to reconstruct the original state space inversely. Therefore, it follows
\begin{equation} \tilde{X}_k=\psi_k\circ \phi_k(X_k) \label{AE}\end{equation}
where $\tilde{X}_k$ are the reconstructed states. However, only the encoder plays a functional part in our framework following Eq. \eqref{encoder}. The decoder ($\psi_k: H_k \rightarrow X_k$) is used for computing a regularization term, called reconstruction loss, to train $\phi_k$:
\begin{equation} \mathcal{L}_{\text{recon},k} = \frac{1}{n} \sum^{n}_{i=1}\| (X_k - \tilde{X}_k)_i \|^2_2 \label{reconloss}\end{equation}
where $n$ is the total number of data points. Note that $\psi_k$ is omitted in Fig. \ref{fig:fig2}.

\subsection{Koopman transition module}
In MMSs, the product quality at stage $S_k$ is determined by the local process characteristics and the quality inherited from stage $S_{k-1}$. Therefore, the general expression of the quality indicators follows
\begin{equation} H_k=f_k(X_k, H_{k-1}) \label{quality}\end{equation}
where $H_k \in \mathbb{R}^{d_{h,k}}$, $f_k$ combines an encoding model and a propagation model. This recursive expression indicates that $H_k$ summarizes all the process measurements prior to $S_k$. Our goal here is to identify the universal linear embedding for a nonlinear system and clearly observe the quality propagation in the latent space. Therefore, we seek Koopman operators that induce the linear aggregation of quality indicators following:
\begin{equation} H_k=\hat{H}_k+\mathcal{K}_{k-1}H_{k-1} \label{qualityind}\end{equation}
In practice, $\mathcal{K}_{k-1}$ can be approximated by a Koopman matrix $\mathbf{K}_{k-1}$. There are several ways to set up the Koopman matrix, including dense matrices \cite{balakrishnan2021stochastic}, or Jordan blocks \cite{lusch2018deep}, etc. In the view of Eq. \eqref{EigenEqn}, we choose to use a diagonal Koopman matrix that consists of the eigenvalues corresponding to the Koopman eigenfunctions identified by AEs. $\mathbf{K}_{k-1}$ constrains a decoupled propagation between the quality indicators from adjacent stages. The method of constructing the matrix is inspired by \cite{lusch2018deep}, as shown in Fig. \ref{fig:fig2}. An auxiliary network is used to parameterize the eigenvalues $\lambda$. The eigenvalues are then aggregated and reshaped to form the diagonal Koopman matrix. It follows $\mathbf{K}_{k-1}=diag(\lambda) \in \mathbb{R}^{d_{h,k-1}\times d_{h,k}}$. 

Unlike conventional Koopman approaches where $\mathbf{K}_{k-1}$ is a static matrix, $\mathbf{K}_{k-1}$ is allowed to vary according to different $X_k$ and $H_k$ in our design. The advantage of this design is: (1) the quality indicators are decoupled from each other, which provides a clear view of the quality evolution; and (2) it allows input-dependent quality propagation, in contrast to the static propagation model in \cite{yan2021deep}, meaning the proposed method can adapt to a wide range of operating conditions without sacrificing accuracy. Even though the input-dependent $\mathbf{K}_{k-1}$ yields only a piecewise (not globally) linear latent system, it allows one to account for nonlinear systems with continuous spectrum using a low-dimension network \cite{lusch2018deep}.

\subsection{Prediction module}
After sequential propagations, a two-layer MLP network is used to predict the quality indices $Y_k$ using $H_k$. To this end, we have introduced all the essential modules to predict the quality indices as in Eq. \eqref{xyCorrelation}. For the deterministic model, it follows:
\begin{equation}
\tilde{Y}_k=\text{MLP}_k(H_k) \label{SimpleMLP}
\end{equation}
where $H_k$ is updated following a linear transition rule as defined in \eqref{qualityind}. Substituting \eqref{qualityind} into \eqref{SimpleMLP} yields:
\begin{equation} \tilde{Y}_k=\text{MLP}_k\left(\sum_{i=1}^{k}\left(\Big(\prod_{j=i}^{k}\mathbf{K}_{j}\Big)\phi_i(X_i)\right)\right) \label{comprehensive}\end{equation}
where $\mathbf{K}_k=I$ for the current stage $S_k$. The expression between $\tilde{Y}_k$ and $X_k$ represents a nonlinear system. However, the cumulative product of $\mathbf{K}_k$ in Eq. \eqref{comprehensive} yields a linear system in the latent space. Our method first selects the most valuable parameters to find the Koopman embedding, which is then combined with MLP to capture the nonlinearity in a linear manner. The Pseudocode of the prediction process can be found as follows.

\begin{algorithm}[H]
\caption{Prediction of quality indices}
\begin{algorithmic}[1]
\Require Process data $X_1,\ldots,X_N$, Quality data $Y_1,\ldots,Y_N$
\While{$k<N$}
\State Compute local latent variable $\hat{H}_k=\phi_k(X_k)$
\If {$k=1$}
\State Compute $H_k=\hat{H}_k$
\Else 
\State Compute cumulated latent variable $H_k=\hat{H}_k + K_{k-1}H_{k-1}$
\EndIf
\State Predict $\tilde{Y}_k=\text{MLP}_k(H_k)$
\State Compare $\tilde{Y}_k$ with $Y_k$ when desired
\EndWhile
\end{algorithmic}
\end{algorithm}

\subsection{Two-step training and loss functions}
Many traditional machine learning algorithms require feature engineering before building prediction models, whereas deep learning models enable combining these two tasks together. In \cite{liu2020adversarial, zhang2021path}, process data are converted into a unified feature space using feedforward neural networks. However, training a DNN with a complicated structure can be unstable. For example, the diffusion of gradients may happen when backpropagating errors to the early stages of an MMS. In practice, we propose a two-step training for our framework. This can be useful when global optimality is difficult to obtain.

\begin{enumerate}
\item \emph{Step 1 (pre-training)}: This step requires the encoding and prediction modules to be trained separately. An AE is trained for each stage to minimize \eqref{reconloss}, allowing for unsupervised feature learning. This will guide the encoder modules to find optimal local latent embeddings that best reconstruct the original data. Then, starting from stage 1, the Koopman transition modules are trained together with their consequent prediction modules. These modules can be trained jointly, with the temporal quality indicators and the propagated quality indicators being inputs, and the ground truth quality measurements being outputs. Fig. \ref{fig:fig3} shows the training process at stage II. The loss function follows:
\begin{equation} \mathcal{L}_{\text{pred},k}=\frac{1}{n} \sum^{n}_{i=1}\| (Y_k - \tilde{Y}_k)_i \|_2^2 \label{predloss}\end{equation}
The performance of the pre-trained model is usually limited because it cannot guarantee that the extracted quality indicators are universally optimized for all stages. Therefore, a fine-tuning step is required.

\begin{figure}
\centerline{\includegraphics[width=0.3\columnwidth]{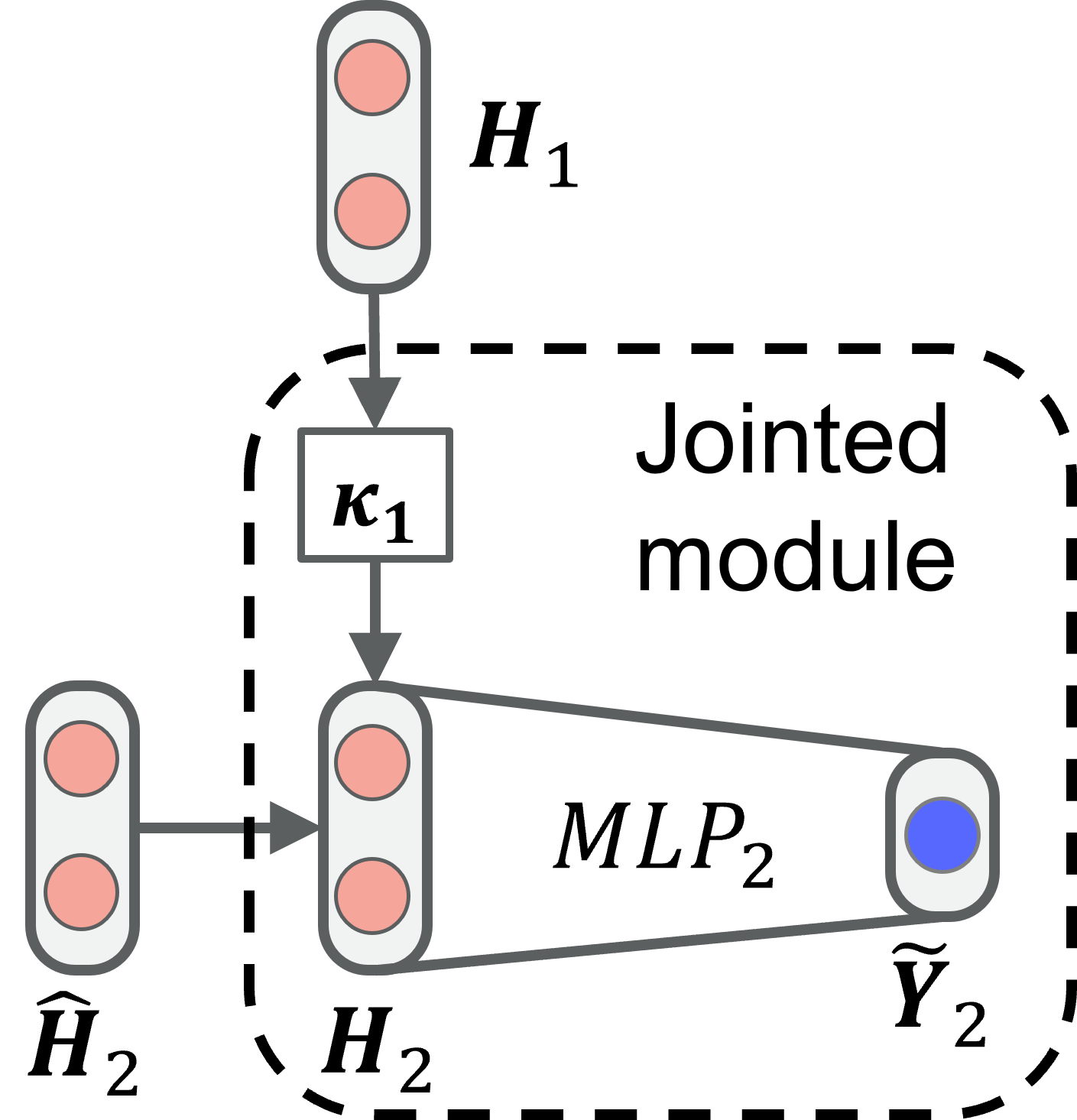}}
\caption{Pre-training of the prediction module at stage II.}
\label{fig:fig3}
\end{figure}

\item \emph{Step 2 (fine-tuning)}: All modules are connected to form a complete network model. The loss function consists of the penalty for prediction errors and other regularization terms. The proposed deep Koopman model must enable quality estimations not only at the final stages, but also at the intermediate stages, and likewise for reconstructing raw data from latent variables. Therefore, we define the full loss function as:
\begin{equation} \mathcal{L}_\text{total}=\sum_{i=1}^{N} \bigg( \rho_i \mathcal{L}_{\text{pred},i} + \theta_i \mathcal{L}_{\text{recon},i} \bigg) \label{totalloss}\end{equation}
where $\rho_i$ and $\theta_i$ define the significance of the prediction and reconstruction at each stage. In our implementation, all $\rho_{i}$ are set to be 1, meaning that all quality indices are equally important. While minimizing $\mathcal{L}_{\text{pred},i}$ is our primary goal, the reconstruction error serves as a regularization term that helps mitigate overfitting.
\end{enumerate}

\emph{Remark 1}: For most MMSs, various operations are conducted at different stages. Therefore, a straightforward linear summation of process variables is unsuitable, as it violates the principles of dimensional analysis. In the proposed method, the autoencoders first transform the process variables into a common feature space, where all latent variables have consistent dimensionality. Therefore, linear propagations can be enforced in the latent space.

\subsection{Stochastic deep Koopman model (SDK)}

\begin{figure}
\centerline{\includegraphics[width=0.9\columnwidth]{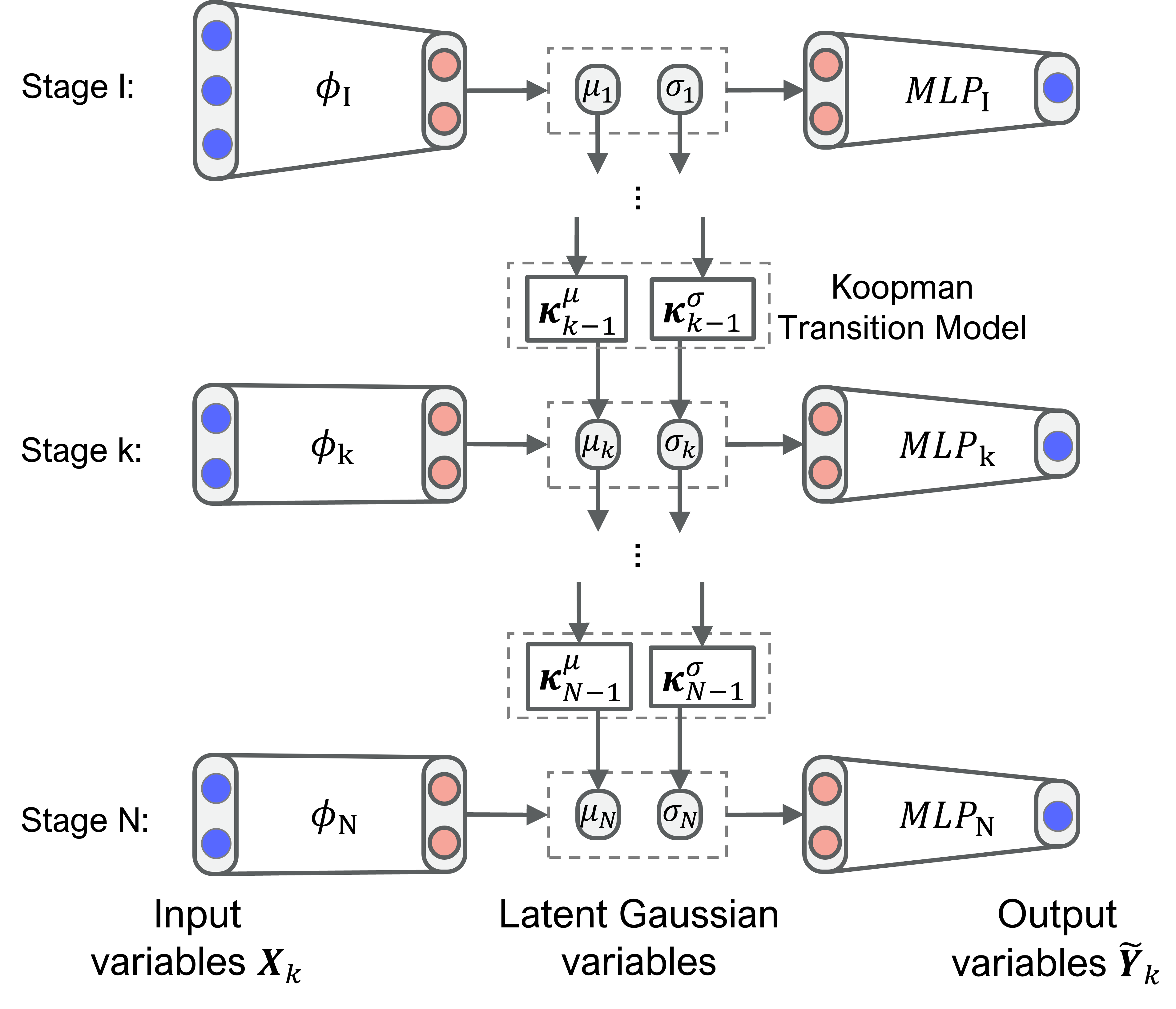}}
\caption{In the stochastic framework, quality indicators are modeled as Gaussian variables.}
\label{fig:fig4}
\end{figure}

In manufacturing systems, the same operation setting usually leads to varying product quality due to various sources of process noise. With such considerations, the base model may not capture the random deviations present in realistic processes. We propose an augmentation that improves the performance of the model to deal with uncertainties. In this stochastic approach, we replace the AE with VAE and model the quality indicators as Gaussian variables. The updated framework is shown in Fig. \ref{fig:fig4}. In the stochastic framework, the encoding module first encodes the state into latent variables $\hat{H}_k$ as in Section \ref{sec: AE}. Then, we learn the Gaussian distribution of $\hat{H}_k$ using two single-layer networks, which give:
\begin{equation} P(\hat{H}_k \mid X_k)\sim\mathcal{N}(\hat{\mu}_k(X_k), \hat{\sigma}^2_k(X_k)) \label{gaussian}\end{equation}
where $\hat{\mu}_k$ is the vector of temporal mean values and $\hat{\sigma}_k$ is the vector of temporal variance. One can refer to \cite{kingma2013auto} for implementation details of VAE. 

Following Eq. \eqref{qualityind}, one can derive a similar expression of the transition of the Gaussian variables. If $P(\hat{H}_k) \sim \mathcal{N}(\hat{\mu}_k, \hat{\sigma}^2_k)$ and $P(H_{k-1}) \sim \mathcal{N}(\mu_{k-1}, \sigma^2_{k-1})$ are independent distributions, then $P(H_k) \sim \mathcal{N}(\mu_{k}, \sigma^2_{k})=\mathcal{N}(\hat{\mu}_k + \mathcal{K}_{k-1}\mu_{k-1}, \hat{\sigma}^2_k + \mathcal{K}_{k-1}\sigma^2_{k})$. However, the distributional dependency is often unclear. Once again, we use Koopman operators to approximate the transition of $P(H_k)$. The propagation of the Gaussian variables is given by \cite{balakrishnan2021stochastic}:
\begin{equation} \mu_k=\hat{\mu}_k+\mathcal{K}^{\mu}_{k-1}\mu_{k-1} \label{KoopmanMean}\end{equation}
\begin{equation} \ln{\sigma_k}=\ln{\hat{\sigma}_k}+\mathcal{K}^{\sigma}_{k-1}\ln{\sigma_{k-1}} \label{KoopmanVariance}\end{equation}
For the case where VAE is applied, the design of Koopman matrices is similar, except that two distinguished auxiliary networks need to be introduced. Outputs from the two networks are used to construct $\mathbf{K}^{\mu}_{k-1}, \mathbf{K}^{\sigma}_{k-1} \in \mathbb{R}^{d_{h,k-1}\times d_{h,k}}$, correspondingly. Once the distribution of the Gaussian variable is obtained, one can use the reparameterization trick to reconstruct exact local quality indicators. That is, generate a random variable $\epsilon \sim \mathcal{N}(0,1)$ to sample $H_k=\mu_k+\epsilon\sigma_k$.

When pre-training the stochastic model, the reconstruction loss will follow the same expression in Eq. \eqref{reconloss}. However, another regularization term needs to be introduced in the training phase to enforce the Gaussian distribution in the latent space:
\begin{equation} \mathcal{L}_{\text{KLD},k} = D_{KL}\left(P(\hat{H}_k \mid X_k) \| \mathcal{N}(0,I)\right) \label{KLD}\end{equation}
where $D_{KL}$ is the Kullback-Leibler divergence and $\mathcal{N}(0,I)$ is the standard normal distribution. To fine-tune the model, the complete loss function becomes:
\begin{equation} \mathcal{L}_\text{total}=\sum_{i=1}^{N} \bigg( \rho_i \mathcal{L}_{\text{pred},i} + \theta_i \big(\mathcal{L}_{\text{recon},i} + \omega_i \mathcal{L}_{\text{KLD},i}\big) \bigg) \label{VAEtotalloss}\end{equation}
where $\omega_i$ is the weight of $\mathcal{L}_{\text{KLD},i}$. 
\section{Case Study: Results and Analysis}
\label{sec: case study}
\begin{figure}
\centerline{\includegraphics[width=\columnwidth]{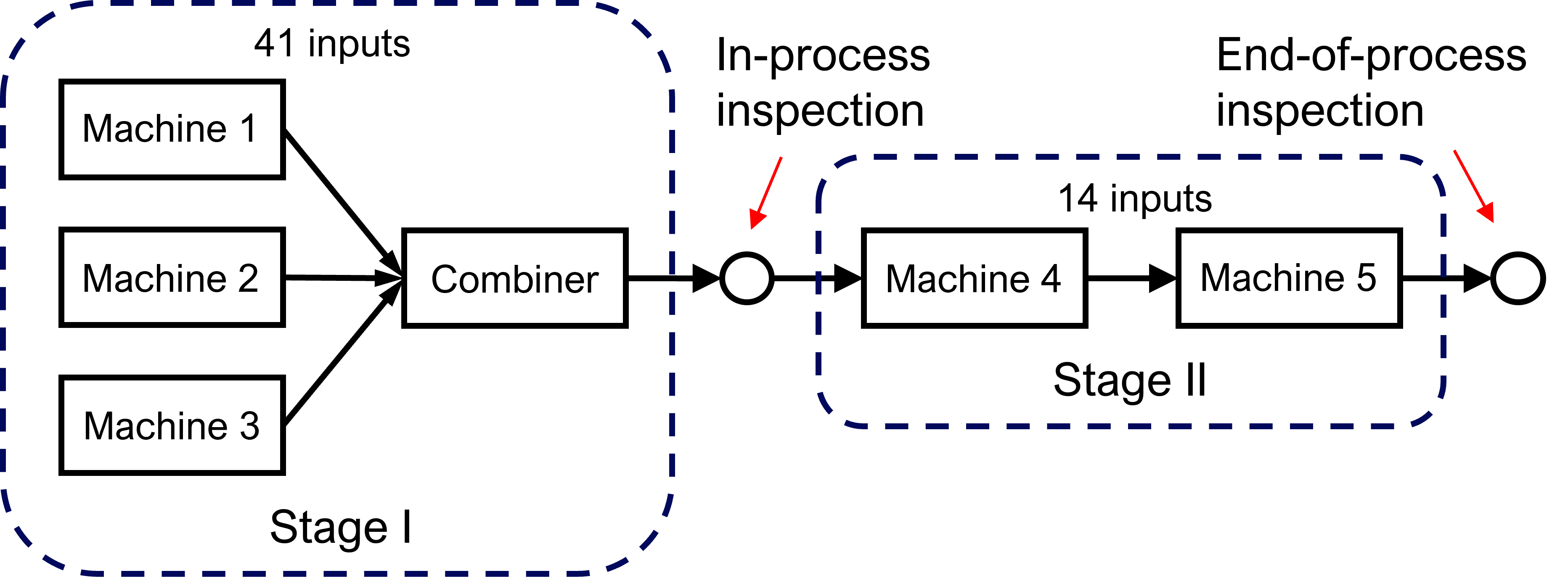}}
\caption{A two-stage manufacturing process with five machines.}
\label{fig:fig5}
\end{figure}

The performance of the proposed method is tested on an open-source public dataset, named ``Multistage continuous-flow manufacturing process'' (MCMP), from Kaggle \cite{kaggle}. The dataset was collected from a high-speed, continuous manufacturing production line near Detroit, MI, USA. Fig. \ref{fig:fig5} shows the configuration of the two-stage-five-machine system. 

There are three identical machines connected in parallel at stage I, whose outputs are merged into a combiner. Then the semi-finished products are further processed by two serial machines at stage II. Operators recorded 12 process characteristics from each machine at stage I, 3 from the combiner, and 7 from each machine at stage II. In addition, there are 2 ambient environmental factors collected during the operation. They are aggregated with the features from stage I. The in-process and end-of-process inspections check the product quality at 15 exact locations of the product. In our implementation, there are 41 features from stage I ($p_1=12\times3+3+2$) and 14 features from stage II ($p_2=7+7$). The 15 quality measures from each stage are regarded as labels. The objective is to use the features to predict the quality labels. From the accompanying information of the dataset, the process measurements are dimensionless, and product qualities are each measured in millimeters.

\subsection{Data Pre-processing}
The quality measurements from MCMP are noisy and error-prone, as discussed in \cite{zhang2021path,kaggle,oleghe2020predictive}. \cite{zhang2021path} devises a masking technique to denoise the labels. However, this masking technique requires manual evaluation to correct labeling errors. We suggest a simple automated way to denoise the labels that significantly reduces the workload of data pre-processing. We first drop all the labels with more than 20\%\ zero-readings from the dataset; then, for most of the remaining attributes, the measurements that are beyond 3.5 standard deviations from the median values are presumed to be outliers and discarded. There are 8 labels remaining for stage I and 13 for stage II after the cleaning, i.e., $q_1=8, q_2=13$.

\subsection{Implementation Details}
We randomly split the MCMP dataset into three parts, i.e., the training set (70\%), validation set (10\%), and testing set (20\%). The architecture of the stochastic Koopman model is selected based on its performance on the validation set. In our implementation, we enforce all stages to share the same latent size, i.e., $d_{h,1}=d_{h,2}=60$. This is reasonable because we create a sufficiently large latent space, and each operation stage only affects a subset of the latent variables. Consequently, we have $\mathcal{K}_1, \mathcal{K}^{\mu}_{1}, \mathcal{K}^{\sigma}_{1} \in \mathbb{R}^{60 \times 60}$. Other hyperparameters include batch size (set to 64), learning rate (set to $10^{-3}$ for the pre-training phase and $3\times 10^{-4}$ for the fine-tuning phase), and constant weights $\theta_1=\theta_2=0.1$, $\omega_1=\omega_2=5\times10^{-5}$. Recall that we define $\rho_1=\rho_2=1$. The algorithm is implemented using the Pytorch library, and the Adam optimizer is used to train the model.

\subsection{Performance Evaluation}
We demonstrate the performance of the proposed stochastic deep Koopman model through a comparison study between different regression models. The benchmark algorithms implemented in this study include:

\begin{enumerate}
\item ANN refers to the fully connected feedforward neural network. A two-layer ANN is implemented with 256 hidden units. The rectified linear unit (ReLU) activation function is chosen for the intermediate layer.
\item Random forest (RF) is an ensemble of decision trees widely used in data challenges. To avoid overfitting, the RF hyperparameters are set to be: number of estimators $=100$, maximum depth $=30$, minimum samples at a leaf (fractional) $=0.01$. 
\item Path enhanced bidirectional graph attention network (PGAT) is a state-of-the-art algorithm for modeling MMSs. One can refer to the paper \cite{zhang2021path} for more details on hyperparameter settings. Note that the authors used the same MCMP dataset to validate the effectiveness of this algorithm in \cite{zhang2021path}.
\item Deep multistage multi-task learning (DMMTL) is a DNN-based method for modeling MMSs developed in \cite{yan2021deep}. It utilizes feedforward NNs to model both the nonlinear state transitions and the emission functions. The sizes of the hidden states are all selected to be 60.
\item Static AE-Koopman (S-AEK) refers to the proposed base model, whose transition models are static dense Koopman matrices.
\item Eigen AE-Koopman (E-AEK) also refers to the proposed base model where auxiliary networks parameterize the diagonal Koopman transition matrices.
\end{enumerate}

Mean squared error (MSE) on $Y$s is used to evaluate the prediction performance of the above algorithms, together with the proposed stochastic Koopman model. During the experiments, ANN and RF use individual models to predict the quality measures from different stages, whereas the others can obtain predictions under a unified framework. The prediction errors on the testing set of MCMP are shown in Table \ref{table2}. The experiments are repeated 10 times with different random seeds to obtain reproducible results. 

\begin{table}
\centering
\setlength{\extrarowheight}{1pt}
\caption{Comparison of Prediction Errors on the test set}
\label{table2}
\setlength{\tabcolsep}{3pt}
\begin{tabular}{p{80pt}  p{110pt}  p{110pt}  p{110pt}}
\toprule
\textbf{Model} & \textbf{Stage I MSE}    & \textbf{Stage II MSE}   & \textbf{Total MSE}  \\
\midrule
ANN                               & 0.0252 $\pm$ 0.0048   & 0.0281 $\pm$ 0.0021   & 0.0270 $\pm$ 0.0023\\
RF                                & 0.0210 $\pm$ 0.0013   & 0.0257 $\pm$ 0.0020   & 0.0239 $\pm$ 0.0014\\
PGAT \cite{zhang2021path}        & 0.0212 $\pm$ 0.0019   & 0.0244 $\pm$ 0.0021   & 0.0231 $\pm$ 0.0017\\
DMMTL \cite{yan2021deep}         & 0.0252 $\pm$ 0.0016   & 0.0272 $\pm$ 0.0012   & 0.0265 $\pm$ 0.0014\\
S-AEK                             & 0.0232 $\pm$ 0.0026   & 0.0284 $\pm$ 0.0013   & 0.0260 $\pm$ 0.0016 \\
E-AEK                             & 0.0225 $\pm$ 0.0017   & 0.0271 $\pm$ 0.0012   & 0.0250 $\pm$ 0.0009 \\
\textbf{SDK}          & \textbf{0.0193 $\pm$ 0.0013}   & \textbf{0.0243 $\pm$ 0.0014}   & \textbf{0.0220 $\pm$ 0.0011} \\
\bottomrule

\end{tabular}
\end{table}

It can be observed that SDK achieves the lowest prediction error for both stages. Our base models have similar performance as DMMTL because these methods use similar network structures. The proposed scheme is different from DMMTL because we regularize the selection of latent variables so that they can be used to reconstruct the process measurements. Moreover, linear quality propagation is enforced between stages. An interesting phenomenon is that RF performs much better than most of the NN-based methods, which can be attributed to the relatively small size of the dataset. In addition, our pre-processing procedure removes most label noises so that the weakness of RF in fitting noisy data is not revealed. The MSEs from this work show different scales from those reported by \cite{zhang2021path}, likely due to the different data pre-processing treatment. Yet, we reach the same conclusion that quality measurements from stage II are harder to predict than those from stage I. This is because stage II quality depends on a plurality of prior operations, whereas stage I is unaffected by subsequent stages. Also, the provided stage II measurements are possibly insufficient for accurate predictions.

\begin{figure}
\centerline{\includegraphics[width=0.8\columnwidth]{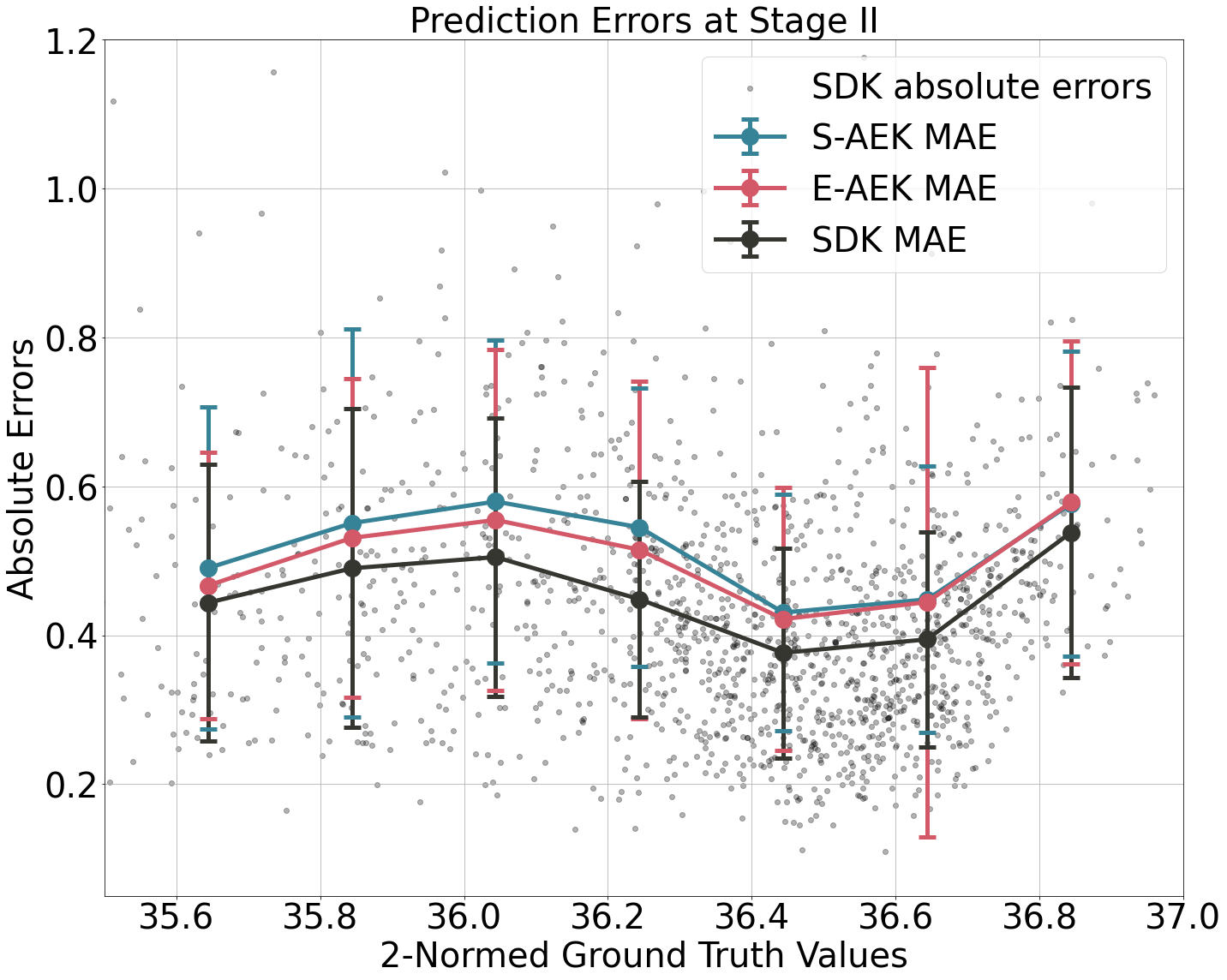}}
\caption{Prediction errors on labels at stage II by different models. The curves illustrate the means and the corresponding error bars computed from data points located in a $\pm$0.1 neighborhood centered around the 2-normed ground truth values.}
\label{fig:fig6}
\end{figure}

We see increasingly better performance when augmenting the base model by introducing the auxiliary network and stochasticity to the latent space. Results from an ablation study that justify these findings are shown in Fig. \ref{fig:fig6}. Notice that for better visualization, mean absolute error (MAE) instead of MSE is shown, and the 2-norm of all labels at stage II ($\|Y_2 \|_2$) is used for the x-axis. S-AEK and E-AEK achieve similar performance in the region $[36.4,36.8]$, which can be considered the quality region under nominal operations. However, their performance deviates when the operating condition starts to shift. E-AEK outperforms S-AEK in the region $[35.6,36.2]$ because the auxiliary network parameterizes an input-dependent transition of the latent variables, enabling coverage of a more extensive range of operating conditions. SDK performs no worse than the other algorithms in all situations because of its robustness in accounting for process noise and outliers.

\subsection{Analysis in the Latent Space}
Operators can perform analytics directly on the latent variables instead of the raw data to understand how quality evolves as the products are transmitted along the production line. 

\begin{figure}
\centerline{\includegraphics[width=0.7\columnwidth]{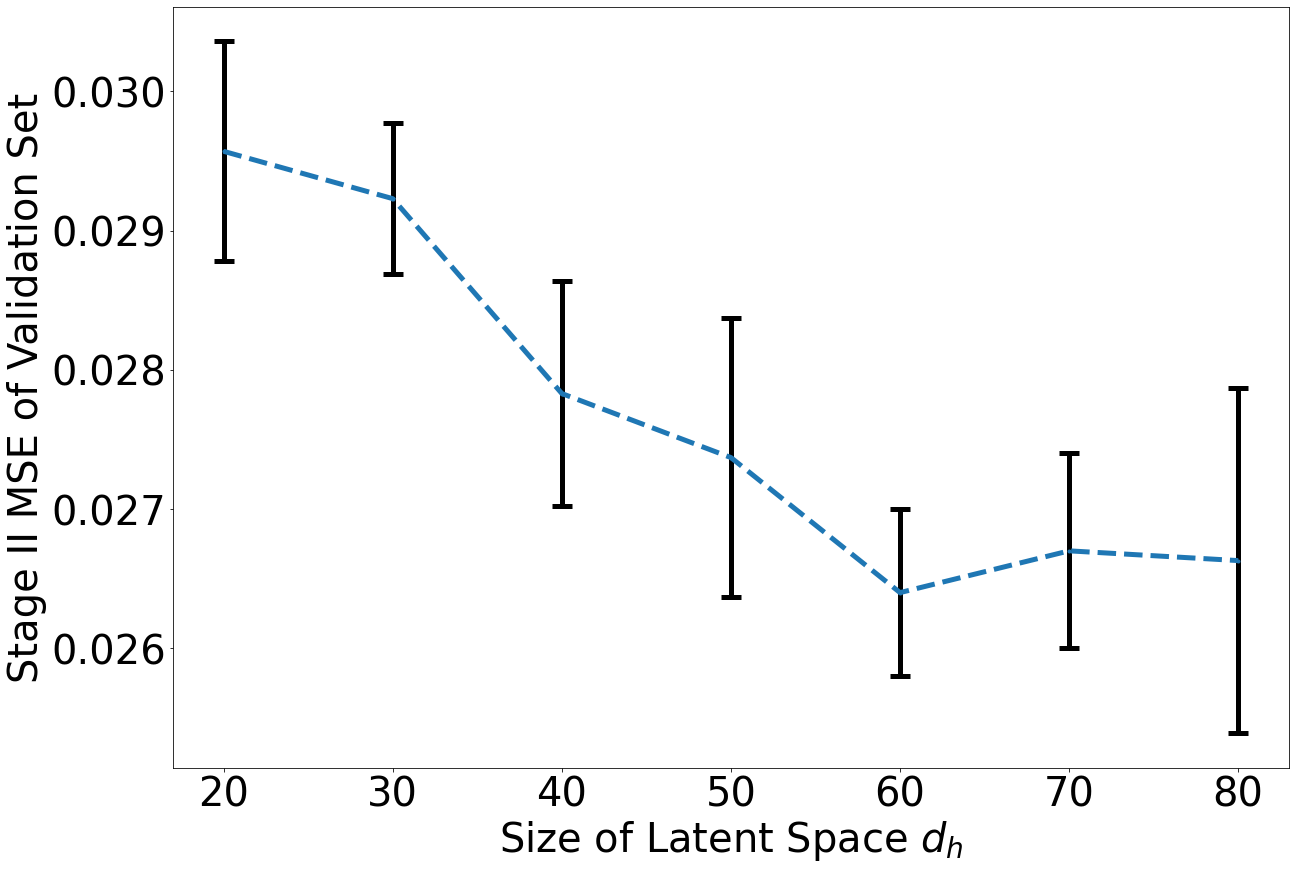}}
\caption{The latent size is decided to be 60 using the elbow method.}
\label{fig:fig7}
\end{figure}

It is critical to decide the latent space dimension. Unlike many other DNN models where autoencoders are used to learn a compact subset of features, the Koopman-based architectures usually require autoencoders to lift the dimension of features so that the linear approximation can be well enforced. However, setting excessively large latent sizes can cause the model to encounter problems such as high computational expense and vanishing gradients. In our work, the latent space dimension is decided to be $d_{h,k}=60$ for all stages based on the Stage II MSE of the validation set, as shown in Fig. \ref{fig:fig7}.

\begin{figure}
\centerline{\includegraphics[width=\columnwidth]{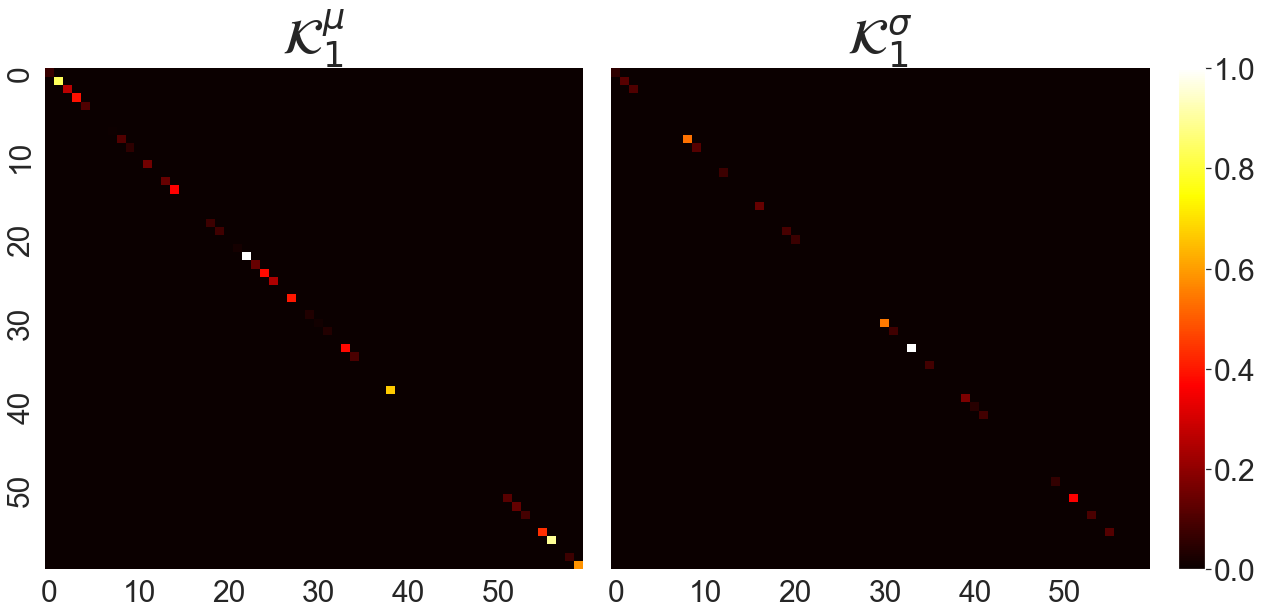}}
\caption{The layout of the Koopman transition matrices under the nominal operating condition.}
\label{fig:fig8}
\end{figure}

Once an SDK (or the base models) is trained, the Koopman transition matrices provide direct visualization of how the critical quality indicators are propagated to the downstream stations, as shown in Fig. \ref{fig:fig8}. The matrices are normalized so that the diagonal terms are scaled within $[0,1]$. It can be observed that $\mathcal{K}_1^{\mu}$ and $\mathcal{K}_1^{\sigma}$ have 35 and 20 non-zero eigenvalues, respectively. The sparsity indicates that around half of the quality indicators from stage I significantly impact the quality at stage II under the nominal operating condition. The rest of the quality indicators will not significantly influence the operation at stage II, but they are critical for performing local quality estimations at stage I. The simple structure of the devised Koopman matrices allows for streamlined process monitoring. For example, operators can quickly locate the root causes by backtracking through the latent space when unexpected quality disruptions occur. We highlight this additional interpretability as an important advantage of our approach over the other benchmarks.

The proposed approach has an inseparable connection with SoV methods. By constructing propagation models, we both attempt to represent product quality indices in terms of process measurements from all prior stages. However, our method resolves some key challenges under the SoV framework while preserving its per-stage structure. In SoV, it is usually assumed that there is prior knowledge of the key process characteristics and propagation mechanisms, which is not always available. In contrast, our method efficiently learns the key quality indicators and Koopman matrices from data. The latent, decoupled transition model in SDK is precisely a variant of SoV. Therefore, SoV-based methods can be adapted to our framework. For example, one can potentially design a linear process controller following \cite{djurdjanovic2017multistage} based on the Koopman models in the latent space.

\begin{figure}
\centerline{\includegraphics[width=0.6\columnwidth]{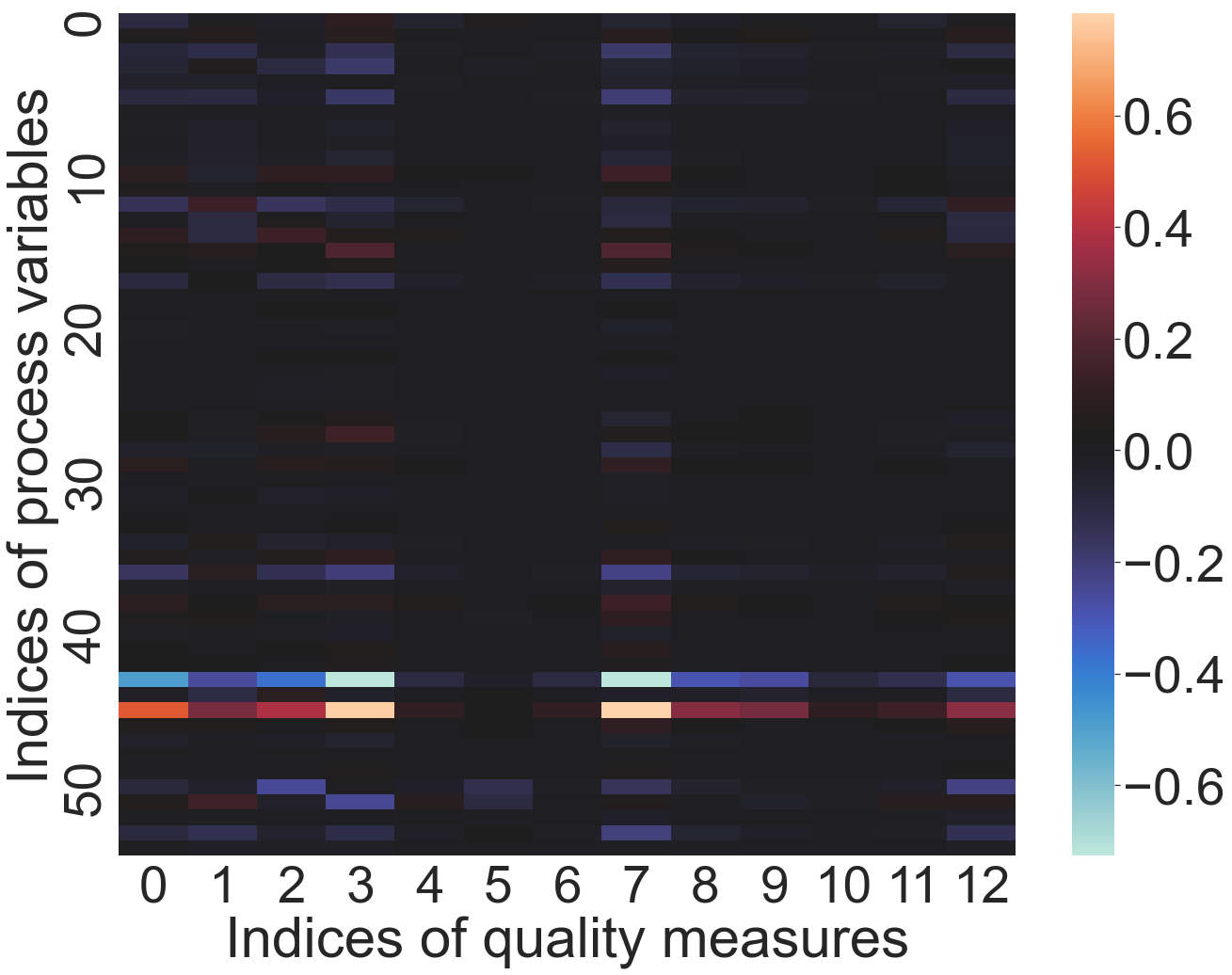}}
\caption{Significance of all input variables w.r.t. final quality indices at stage II, computed by sensitivity analysis.}
\label{fig:fig9}
\end{figure}

\subsection{Interpretability of SDK}
The quality propagation in the latent space of SDK provides a certain level of interpretability by its nature, as the Koopman modules explicitly demonstrate the linear transitions. However, the encoding and prediction modules are neural network models that are lacking in explainability. There have been extensive studies on improving the interpretability of DNN-based models \cite{saltelli2010variance, zurada1994sensitivity}. The most common approach is to perform gradient-based sensitivity analysis to decide the significance of input variables. Note that the computation of sensitivity for a stochastic network like SDK is tricky due to the reparameterization technique. Here, we compute the quality-to-input sensitivity under the nominal operating condition by setting the random variable $\epsilon$ to zero when regenerating $H_k$ from Gaussian variables. In Fig. \ref{fig:fig9}, a brighter element indicates a stronger sensitivity between the corresponding quality index and process variable. The sensitivity shows that a sparse subset of the process characteristics mainly determines the final quality indices. Some quality indices are loosely correlated with the process variables, indicating that more process information needs to be recorded in the future for better monitoring of the process.

\subsection{Discussion}
One of the primary reasons most existing VM architectures are unsuitable for MMSs is their reliance on splicing multiple models together to account for interstage quality transitions in MMSs. In contrast, SDK utilizes a comprehensive model to perform stage-by-stage quality estimations. By minimizing the multi-objective loss function in Eq. \eqref{VAEtotalloss}, the intermediate quality indicators can be better coordinated to ensure high prediction accuracy for all stages. The SDK framework also facilitates the design of root cause analysis and real-time control algorithms. Through sensitivity analysis, we have demonstrated its capability to pinpoint critical variables that may cause potential quality disruptions, significantly expediting the decision-making process for appropriate adjustments.

It is important to acknowledge certain limitations of the SDK algorithm. Firstly, the algorithm necessitates a substantial amount of industrial data for training. Deploying SDK requires investing in additional sensors to collect process measurements. This increases the upfront cost and extends the setup timeline. Additionally, the proposed framework yields a lifted latent space and a two-step training scheme, which together raise the concern of increased computational complexity. However, based on our test, the overall training time of our algorithm remains comparable to the NN-based benchmarks we tested against. To summarize, while the practitioners should be aware of the potential financial, time-related, and computational costs associated with deploying the algorithm, these initial investments should be justified by improved operational efficiency of the system.
\section{Conclusion}
\label{sec: conclusion}
In today's competitive business environment, achieving stable product quality is crucial for reducing production costs and contributing to sustainability. In line with this objective, the concepts of ZDM suggest the development of advanced defect detection schemes for manufacturing systems, with VM techniques serving as a key enabler to meet this need.

The paper presents a stochastic deep Koopman model to estimate quality indices during production, assisting real-time defect detection in MMSs. The performance of the proposed method is validated by a case study on an open-source dataset. The specific conclusions are:
\begin{enumerate}
\item A novel quality estimation framework is presented for MMSs. By combining variational autoencoders and Koopman operators, the SDK model captures the quality propagations along the production line and provides quality estimations in a stage-by-stage manner.
\item The SDK model exhibits improved interpretability compared to conventional data-driven algorithms. It enhances the traceability of quality variations and facilitates the design of root cause analysis schemes, thereby enabling effective quality control.
\item The proposed SDK framework serves as a versatile VM tool for MMSs. Leveraging modern deep learning techniques, it requires minimal knowledge of the underlying physics of the production line. Therefore, it can potentially be adapted to a wide range of manufacturing use cases.
\end{enumerate}

There are several directions for future research that can build upon this work. Firstly, one potential approach to address the data acquisition issue is to integrate federated learning techniques \cite{yang2019federated}, which enables collaborative model development \cite{mehta2022federated}. Secondly, to further enhance the model performance, physical knowledge can be incorporated into the proposed data-driven model, for example, through the techniques from \cite{raissi2019physics}.

More broadly, future work can concentrate on extending the algorithm to cover the complete spectrum of quality control aspects in the ZDM paradigm. Real-time quality control schemes can be developed for complex MMSs based on the proposed framework, as outlined in \cite{chen2023control}. It is also important to investigate the economic and sustainable benefits of the proposed framework. Conducting cost-effective analysis for MMSs following established roadmaps in \cite{tayyab2020sustainable, kugele2023reducing} will allow operators to better understand the return on investment and make informed decisions.

By addressing these research directions, we can advance the application of the proposed algorithm, improve its practicality, and unlock its potential for enhancing sustainability in manufacturing operations.
\section*{Acknowledgment}
This work was financially supported by Ford Motor Company under Award Number 000106-UM0287. The authors would like to acknowledge the collaboration which made this research possible.

\bibliography{reference}
\bibliographystyle{abbrv}

\end{document}